\documentclass[sigconf]{acmart}                
\hypersetup{draft}

\AtBeginDocument{%
  }

\setcopyright{acmcopyright}
\copyrightyear{2023}
\acmYear{2023}
\acmDOI{10.1145/3545YYY.YYYYYYY}

\acmConference[ACMSE 2023]{2023 ACM Southeast Conference}{April 12--14, 2023}{Virtual Event, USA}
\acmBooktitle{2023 ACM Southeast Conference (ACMSE 2023), April 12--14, 2023, Virtual Event, USA}
\acmPrice{15.00}
\acmISBN{978-1-4503-9921-0/23/04}

\setcopyright{none}
\renewcommand\footnotetextcopyrightpermission[1]{} 
\begin{document}
\pagestyle{empty}

\title{A Survey on Conversational Search and Applications in Biomedicine}

%
\author{Naga Sai Krishna Adatrao}
\affiliation{%
  \institution{Kennesaw State University}
  \city{Kennesaw}
  \state{Georgia}
  \country{USA}}
\email{nadatrao@students.kennesaw.edu}

\author{Gowtham Reddy Gadireddy}
\affiliation{%
  \institution{Kennesaw State University}
  \city{Kennesaw}
  \state{Georgia}
  \country{USA}}
\email{ggadired@students.kennesaw.edu}

\author{Jiho Noh}
\affiliation{%
  \institution{Kennesaw State University}
  \city{Kennesaw}
  \state{Georgia}
  \country{USA}}
\email{jnoh3@kennesaw.edu}

\renewcommand{\shortauthors}{Adatrao and Gadireddy, et al.}

\begin{abstract}
This paper aims to provide a radical rundown on Conversation Search (ConvSearch), an approach to enhance the information retrieval method where users engage in a dialogue for the information-seeking tasks. In this survey, we predominantly focused on the human interactive characteristics of the ConvSearch systems, highlighting the operations of the action modules, likely the Retrieval system, Question-Answering, and Recommender system. We labeled various ConvSearch research problems in knowledge bases, natural language processing, and dialogue management systems along with the action modules. We further categorized the framework to ConvSearch and the application is directed towards biomedical and healthcare fields for the utilization of clinical social technology. Finally, we conclude by talking through the challenges and issues of ConvSearch, particularly in Bio-Medicine. Our main aim is to provide an integrated and unified vision of the ConvSearch components from different fields, which benefit the information-seeking process in healthcare systems.
\end{abstract}






\maketitle

\section{Introduction}

\subsection{Information-Seeking Process}

The information-seeking process via human-computer interactions has become more ``interactive'' than ever before. For instance, modern web search engines provide features like \textit{query suggestions} that attempt to specify the user's original search intent, \textit{featured snippets} provide relevant text snippets that may directly answer the user's query, and \textit{knowledge panels} that presents pertinent information to a query in a structured format. According to a market research company~\citep{pricewaterhousecoopers}, 65\% of middle-aged adults speak to their voice-enabled devices at least once a day, which requires conversational features. Designing the features of \textbf{Conversational Search (ConvSearch)} has recently become a more critical factor in developing information retrieval (IR) systems and advances in natural language understanding.

The lexical definition of \textit{conversation} is the act of interchanging thoughts, information, etc., by spoken words between multiple parties. We apply the exact definition to the ConvSearch systems and focus on the properties of conversation that include (1) the use of natural language, (2) multi-turn interactions, and (3) the aspect of the information-seeking process. 

\subsection{History of Conversational Search}

The idea of designing a search engine that allows users or the system to interact with the other party through dialog to seek users' information needs is not a new thing. \citet{Croft1987I3R} stated that ``the IR system should enter into the dialogue with a user to verify inferred concepts and request more information on domains that are not well specified.'' From the late 1990s, researchers studied concrete methodologies to implement dialogue components for IR~\citep{Belkin1995CasesSA} and recommendation systems~\citep{Gker2000TheAP}. Recently, \citet{Croft2019TheIO} stresses on the importance of user-system interaction in the information-seeking scenario, particularly in the limited-bandwidth environments provided by mobile devices and voice-based assistants. 

The volume of scientific literature on ConvSearch has grown exponentially in the last two years (2021-22). Research topics expanded towards multi-modality search~\citep{Nie2021ConversationalIS} and conversational agents for information-seeking in pharmacologic knowledge bases~\citep{Preininger2021DifferencesII}, and so on.

\section{Background}

In the frame of the information-seeking process, IR is closely related to recommender systems and question-answering systems. This section discusses the shared objectives and differences between these related research topics. 

\subsection{Information Retrieval}

Conventional IR systems evaluate ad-hoc queries with documents based on a lexical exact-match model such as BM25~\citep{Robertson2009ThePR} and query likelihood models. Typically, a user specifies the information need with a single query that initiates a search. Then the user receives the system-computed list of relevant items and determines whether to accept the results or refine the initial query for a follow-up search. That sequence completes one session of the information retrieval process.

ConvSearch differs from conventional IR systems in their interaction mode; ConvSearch allows users to describe their information needs in a less strict natural language format. During the conversation, the system can request for additional information from the user when the system's understanding of the initial query is \textit{uncertain}. The effort to communicate back and forth to reduce the uncertainty characterizes the conversational search systems.

\subsection{Recommender Systems}

Recommender systems (RecSys) qualify available items (e.g., products, documents) and make suggestions that may suit the user's expectations or information needs. A recommender system computationally models users' expectations, preferences, item features, and interactions between users and items. Unlike IR, RecSys does not require a user query; suggestions can be made based on item features, user profiles, or interactions.

Recently, conversational approaches to recommender systems (ConvRecSys) allow personalized recommendations through natural language dialog with users, which has become a trending research topic. Notably, the research focuses include (1) refining a user's preference by clarifying preferred item features via conversation~\citep{Zhao2013InteractiveCF,Wu2019DeepLC,Zou2020TowardsQR} and (2) natural language understanding and generation in a conversational RecSys context~\citep{Sun2018ConversationalRS,Liu2020TowardsCR}.

\subsection{Question-Answering}

Within the fields of IR and NLP, question-answering (QA) is the task of computationally answering questions posed by humans in natural language. Like ConvRecSys, Conversational Question-Answering (ConvQA) emphasizes the multi-turn dialogue interactions between users and the system.

Due to the shared objectives between QA and IR, many QA techniques and benchmarks are adapted in the IR research; both systems are expected to respond to the user's information needs. For example, differing from IR, QA seeks to find relevant passages from a document collection or infer over a knowledge graph to answer a question. On the other hand, IR considers a set of relevant documents an answer to the user's query.

ConvQA takes a more vital role in ConvSearch because we aim to provide integrated and dynamic system responses in ConvSearch. In that scenario, a direct answer to a user's query (e.g., a factoid question) is a desirable system response.

\subsection{Scope of Survey}

\subsubsection{Interactive Information Retrieval}

ConvSearch is interactive information retrieval (IIR) because of the mixed-initiative communication between the user and the system~\citep{Radlinski2017ATF}. From the Human-Computer Interaction (HCI) perspective, the medium of IIR communication can be non-verbal (e.g., gestures as sensory or images/videos as visual) or verbal (e.g., spoken as audio or written as textual). We consider speech a verbal communication that requires recognition and translation into text~\citep{Crestani2002SpokenQP,MorenoDaniel2007SpokenQP} for text-based information retrieval systems. This survey will mainly focus on verbal communication, as \citet{Anand2019ConversationalS} described a conversational search system as an IIR system with speech and language processing capabilities. According to human principles, the interactions should be developed on the emotion of the dialogue. Hence, the challenge is to recognize the emotion behind the state of dialogue the user drops, especially in bio-conversational search systems. The intent of the patient is necessary to be taken into consideration. \citet{9615035} illustrated the Conversational Emotion Recognition that can be achieved by self-attention mechanism with an external knowledge base.

\subsubsection{Task-Oriented Conversation}

ConvSearch is a task-oriented dialogue system where the task is specific to IR. So, ConvSearch has the inherent problems of dialogue systems, such as dealing with phenomena like anaphora (use of a word whose interpretation depends upon another expression in context) and ellipsis (omission or suppression of words that appeared earlier) \citep{Walker1990MixedII} or modeling and controlling dialogue structure~\citep{Churcher1997DialogueMS}.

This survey highlights the research efforts on dialogue systems with IR capabilities. However, we do not exclude the approaches toward questions-answering and recommendation tasks.

\subsubsection{Inclusion and Exclusion}

This paper surveys recent advances in ConvSearch technologies and their applications in the biomedical domain. Section~\ref{sec:isp} portrays ConvSearch as an elaboration process in the information-seeking process, which attempts to maximize the certainty to users' information needs. In Section~\ref{sec:cs}, we dissect conversational search systems and present representative research efforts on each component of the conceptual architecture, such as conversational QA, recommender systems, and knowledge bases. In Section~\ref{sec:bio}, we demonstrate information-seeking applications in the biomedical domain. In addition, we discuss the aims of biomedical conversational search systems, likely semantic representations, trustworthiness, and personalization issues.  

We identified relevant works by first querying several scholarly document search systems (i.e., Google Scholar, Semantic Scholar, arXiv, Citeseer, and ResearchGate) with commonly used keywords for conversational search and related research areas, such as ``conversational search,'' ``ConvSearch,'' ``conversational information retrieval,'' and ``CIR search.''  For biomedical applications, we used keywords such as ``biomedical information retrieval'' and ``biomedical document search.'' Further, we expanded the collection by manually selecting representative and relevant papers cited from the documents we have already collected. We have collected 346 citations, and approximately half of the documents are considered in this study for review.

\begin{table*}[tbh]
    \centering
    \begin{tabular}{@{}lllll@{}}
        \toprule 
        & \textbf{I. Initiation} & \textbf{II. Understanding} & \textbf{III. Elaboration} & \textbf{IV. Presentation} \\
        \midrule
        \begin{tabular}{@{}l@{}}Means of\\ communication\end{tabular} & 
            \begin{tabular}{@{}l@{}}
            - search box\\
            - hierarchical trees\\
            - form-based interface\\
            - chatbox
            \end{tabular} &
            \begin{tabular}{@{}l@{}}
            - initial user query\\
            - system response
            \end{tabular} &
            \begin{tabular}{@{}l@{}}
            - query expansion\\
            - relevance feedback\\
            - visualization\\
            - \textbf{conversation}
            \end{tabular} &
            \begin{tabular}{@{}l@{}}
            - retrieval results\\
            - answers to questions\\
            - recommendations
            \end{tabular} \\
        \midrule
        Interactions & & \multicolumn{1}{c}{$Q_0$ and $A_1$} & 
                         \multicolumn{1}{c}{$R_1, A_2 R_2, \ldots, A_{k-1} R_{k-1}$} & 
                         \multicolumn{1}{c}{$A_k$} \\
        \midrule
        \begin{tabular}{@{}l@{}} Understanding\\ user intents\end{tabular} &
        \multicolumn{4}{c}{uncertain \hspace{3mm} $ \xrightarrow{\hspace*{8cm}} $ \hspace{3mm} certain} \\
        \bottomrule
    \end{tabular}
    \caption{Information seeking process~\label{tbl:info-seeking}: Conversation as a mean of clarifying user's information need and adding details to the query.}
\end{table*}

\section{Information Seeking Process~\label{sec:isp}}

This section contextualizes the role of conversation in the theme of the information-seeking process. Table~\ref{tbl:info-seeking} presents the staged process of information-seeking activity in particular to information retrieval systems. We adopt a model of information-seeking behavior proposed by \citet{kuhlthau2004seeking}. Her model comprise six stages: initiation, selection, exploration, formulation, collection, and presentation. From the information retrieval perspectives, we modified it to a simple four-staged process: initiation, mutual understanding, elaboration, and presentation. The following subsections explain each stage and demonstrate examples.

\subsection{Initiation}

In the \textit{initiation} stage, a user begins a search process by recognizing the means of communication for their information-seeking activities. The user interface plays a vital role in this stage and the following \textit{presentation} stage as well. First, interface recognition aids in the users' expression of their information needs. Users formulate a query, select appropriate information sources, filter results by specific options using facets, and later review the search results in the presentation stage.

A standard interface for search engines is an ad-hoc search box that allows a single query string from a user. Typically, users write a short sequence of keywords that describes their information needs. The system may provide query suggestions in this search box, but we consider it as an \textit{elaboration} action in the following stage. 

The emergence of conversational systems demands a different mode of communication, such as voice user interfaces (VUIs) or chatbots~\citep{Gorin1997HowMI,McTear2016TheRO}. As we defined the two characteristics of ConvSearch, this conversational interface has two attributes: the use of natural language and multi-turn interactions.

\subsection{Mutual Understanding}

Provided with the interface features, a user and the system first attempt to explore the topic of interest in the search space. In this stage, a user specifies the information needs, and the system reveals what is available in the data collection. We assume that users may not be familiar with the searching task and even unclear about the target information due to the misconception of the topics or the data collection itself. Depending on the search system, users endeavor to find the best query by which the system can retrieve relevant data and satisfy the user. The system analyzes the initial query utilizing various NLP and IR techniques such as query prediction~\citep{Roy2019EstimatingGM} and query expansion~\citep{Nogueira2019DocumentEB}. Then, the system can ask for clarification or return the initial retrieval results upon the analysis.

\subsection{Elaboration}

Elaboration is an iterative process, often illustrated as a loop between a user and the system in IR process diagrams. The outcome of the interaction is a refined query. We narrow down the search focus through the interactions between a user and the system in this stage. We add additional information to a user query (e.g., query expansion) or reformulate a query given by the user's feedback (e.g., explicit/implicit relevance feedback). We also consider interactions using visual aids as an elaboration action.  We aim to leverage \textbf{conversations} for the elaboration task toward conversational search systems. 

\subsection{Presentation}

The presentation stage is highly linked with communication and the initiation stage. The scope of user interfaces for search has been expanded recently, including textual and audio/video communications, which requires the system to accommodate different interfaces, such as multi-modal (e.g., a conversation using visual information), multi-dimensional (e.g., knowledge panel), and interactive (e.g., visual analytics), including conversational agents (e.g., chatbots).

The content of search responses should also contain answers to questions and recommendations. As the search systems become more flexible with their communication strategies, systems response should include the retrieval results and other types of interactions. This demand creates another set of challenges in search system design. 

\section{Conversational Search~\label{sec:cs}}

In many real-world information-seeking scenarios, seekers lack detailed knowledge of the underlying information sources. Consequently, a query formulation effort (``an intelligent guess'') primarily relies on the user's prior knowledge. However, ConvSearch changes this scenario according to the initiatives and user intents to make information more accessible and satisfying in retrieval practice~\citep{Zamani2022ConversationalIS}. 

Suppose the goal of ConvSearch aims at satisfying a user's information needs in a sequence ($S$) of multi-turn dialogue between the system and a user. 
$$S = Q_0, (A_1 R_1, A_2 R_2, \ldots, A_{k-1} R_{k-1}), A_k$$
A conversation is initiated by a user's first attempt to express his/her information needs ($Q_0$); This can be a keyword-based query or a complete interrogative sentence. The conversation continues with multiple pairs of a system-oriented action ($A_i$) and the user's response ($R_i$). The goal of ConvSearch is to satisfy the user with $(A_k | A_1, \ldots, A_{k-1})$ and ideally terminate the conversation when the user is satisfied with $A_k$.

We do not strictly define or categorize system actions ($A_i$). Still, we include at least the following action types: $A_i \in \{ D_R, D_S, q, r\},$ where $D_R$ is a set of relevant data items, $D_S$ is a recommendation with suggested items, $q$ is a question for search-oriented clarification, and $r$ is pertinent information to any of the previous user-initiated questions (e.g., a direct answer or a textual snippet that can contain an answer to $R_{<i}$). The user can provide further information ($r$) to a question or take the initiative to ask a follow-up question ($q$): $R_i \in \{q, r\}$

\begin{figure*}[tbh]
    \centering
    \includegraphics[width=.8\linewidth]{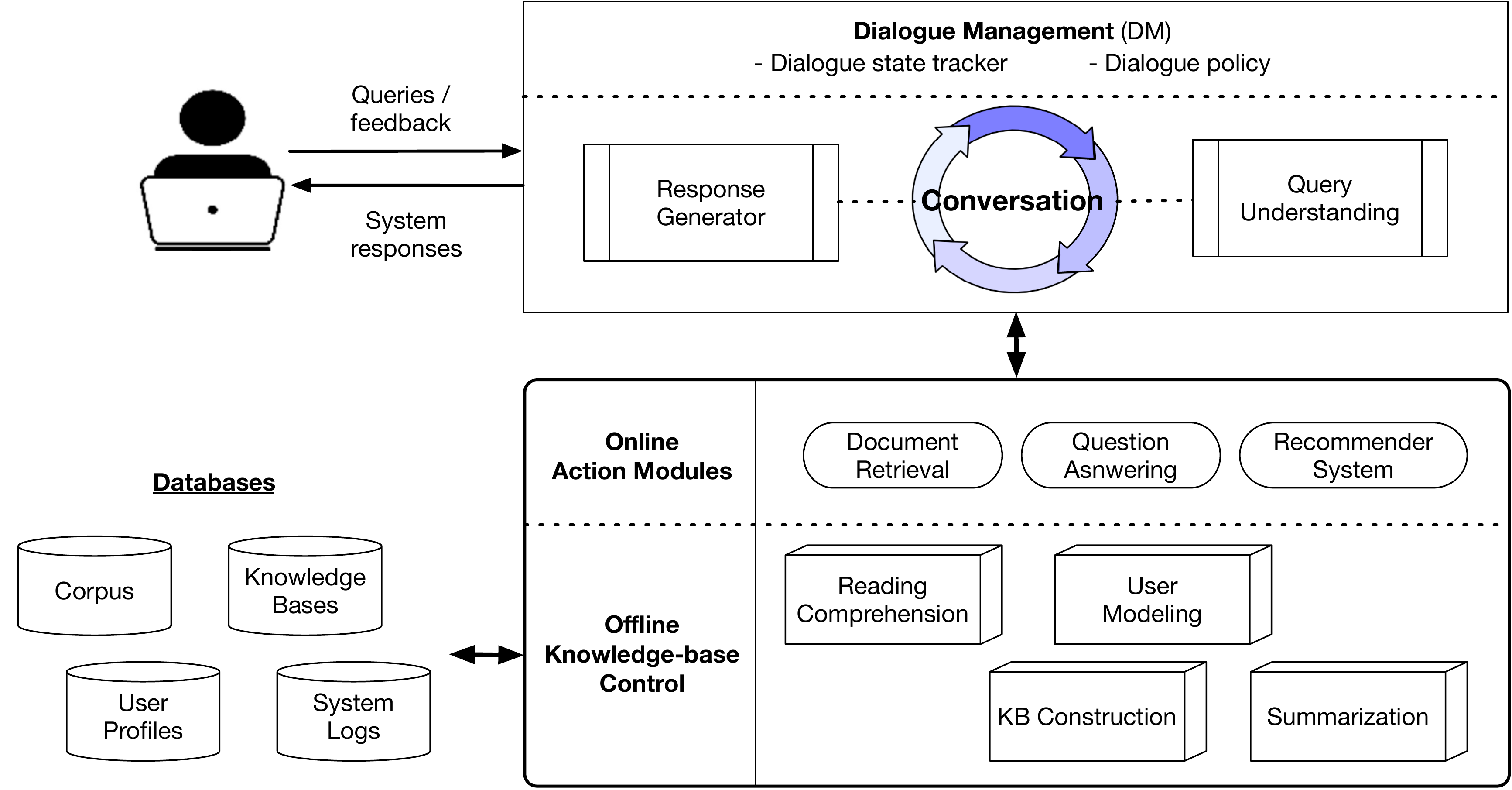}
    \caption{The architecture of a conversational search system: Dialogue Management System (DMS) coordinates the conversation between the system and user. Online Action Modules operate an action requested by DMS and Offline KB Modules prepare information to support the online system actions.}
    \label{fig:cir}
\end{figure*}

Figure~\ref{fig:cir} illustrates the architecture of a ConvSearch system, which consists of three components: (1) a dialogue management system (DMS), (2) online action modules, and (3) offline knowledge-base modules. DMS coordinates the conversation between the system and the user. The two most essential features of DMS are \textit{query understanding} and \textit{response generation}. For example. DMS might involve in a process for specifying the user's intent by generating a clarifying question. The query understanding module analyzes the initial retrieval results and the user query to predict if it is necessary to ask a clarifying question. The response generation module determines how to form a question and executes it.

Online action modules operate a system action requested by DMS, which consists of three modules: information retrieval system, question-answering, and recommender system. ConvSearch systems allow queries in a more natural format such as a complete question sentence, which makes the question-answering feature a crucial module in developing ConvSearch systems. Furthermore, the ConvSearch system should also work as a recommender system to let users explore available data items present in the database and possible search options. We conceive an architecture of ConvSearch systems which comprises two important system action modules and the information retrieval system.

\subsection{Dialogue Management System}

Dialogue Management Systems (DMS) engage in conversational interactions between a user and the system, primarily concerned with understanding a user's intent and generating an appropriate system action to it. Typically, DMS consists of two modules: \textit{dialogue state tracking} and \textit{dialogue policy}. 

With the IR-oriented DMS, the retrieval task is tracked in the form of the belief state to represent the user's information needs throughout the conversation. Methods utilizing structured states, such as frame-based slot filling approaches~\citep{Goddeau1996AFD}, are widely adopted in task-oriented dialogue systems. Unfortunately, this pre-defined slot/value pair structure is considered unsuitable for open-domain conversational search systems~\citep{Cohen2020BackTT}. 


One of the main challenges of \textit{dialogue state tracking} (DST) is identifying entities across a multi-turn conversation, which consequently poses a conversational query reformulation (CQR) problem such as coreference resolution and query expansion~\citep{Rastogi2019ScalingMD}. Researchers attempted to resolve the ambiguity issues in a local query by adding keywords from its session (including answers to any queries) or prior knowledge~\citep{Qu2019BERTWH,Yang2019QueryAA}.  Recently, leveraging neural networks for predicting slot values has been a general approach to the DST problems~\citep{Mrksic2015MultidomainDS,Heck2020TripPyAT,Balaraman2021RecentNM}.

\textit{Dialogue policy} (DP) is a function that decides how the system responds to the user's previous utterances. Rather strict belief state representation is not suitable for the ConvSearch systems. Hence, DP is expected to be a more versatile and flexible learning system with a high priority because the ConvSearch systems integrate subsystems in IR/QA/RecSys/KB/NLP.

In general, DP should select the most effective action to minimize uncertainty, which is the ultimate goal of the ConvSearch systems. More specifically, \citet{Penha2019IntroducingMA} listed fifteen sub-goals (e.g., query disambiguation from IR, intent/domain prediction from DMS, MRC from NLP, etc.) and categorized them into two groups: information-need elucidation and information presentation, which correspond to our last two stages in the information-seeking pipeline (i.e., elaboration and presentation). The action set has no standardized scheme. For example, depending on the search goals, we can suggest actions for each stage as follows: \{\textit{clarify, elicit, interpret}\} for elaboration and \{\textit{provide results, recommend, summarize, explain}\} for presentation.

\textit{Query Performance Prediction} (QPP) is one of the IR evaluation methods actively studied over the past two decades. QPP estimates the effectiveness of the retrieval results without using human relevance judgments. QPP has become more critical in the context of ConvSearch because it can be a significant factor in dialogue policy learning~\citep{Roitman2019ASO}. \citet{Pal2021EffectiveQF} proposed a method that involves shifting a window of terms through a conversation and score the segments using QPP. Recent studies focus on using neural approaches such as using a pre-trained language model for QPP~\citep{Arabzadeh2021BERTQPPCP}.

\subsubsection{Conversational Query Understanding (CQU)}

A query understanding module is expected to transform textual data (i.e., query) into semantic representations by inferring the user's intent using various NLP techniques, including deep learning methods. When it becomes Conversational Query Understanding (CQU), we can formulate this problem as context-aware query reformulation. Essentially, the goal is to construct a conversational query that precisely encodes the user's information needs. Researchers have made significant progress on related tasks, including user intent prediction and query disambiguation. 

The challenges in CQU originated in the unclearness of user utterances due to anaphora and ellipsis. Hence, the query reformulation process, such as coreference resolution and query expansion, is a necessary feature regardless of the type and modality of the search. We consider these problems critical sub-tasks of dialogue management systems. Recently proposed methods commonly leverage neural network models to accomplish these tasks. \citet{Yu2019WhatYS} proposed a visual-aware pronoun coreference resolution method between a dialogue and an image.  \citet{lin2021multi} enhanced a neural query rewriting module using term importance estimation for conversational passage retrieval tasks. 

\subsubsection{System Response Generation}

\textit{Response Generator} elicits outputs by converting structured data into natural language utterances, representing NLG task engines. The output should contain various linguistic components, such as function words and punctuations. \citet{gatt2018survey} categorized the linguistic realization into three different approaches: (1) human-crafted templates (2) human-crafted grammar-based systems, and (3) corpus-based statistical approaches. 

\citet{Mikolov2010RecurrentNN} introduced the recurrent neural network approach for training a language model. A similar approach was advanced by \citet{Wen2015StochasticLG,Wen2015SemanticallyCL} with an LSTM-based structure to NLG. Subsequently, according to the dialogue policy, the challenges in NLG (e.g., information omitting and duplicating problems in surface realization) are addressed by gating the input vector of LSTM-based structure \citep{Wen2015SemanticallyCL,Tran2017SemanticRG}. Recently, \citet{Dudy2021RefocusingOR} highlighted the relevance as in IR to be a crucial measure for designing text-generating tasks.



\subsection{Knowledge Base}

Designing and constructing a knowledge base (KB) is crucial in developing a ConvSearch system. KB represents a knowledge model of the search system, which may integrate different elements such as words, concepts, or entities, and their predicates. Typically, KBs are modeled employing taxonomy or ontology, structured in relational databases or knowledge graphs (KG). Recent research studies explore the efficacy of using KBs embedded in a semantic vector space using neural networks. The following highlights some of the use cases of various knowledge models.


The most dominant use case of KB in the information-seeking process is question-answering. KBQA (QA over KB) is to elicit answers to given questions using the semi-structured relationships between entities, and this field has been studied actively~\citep{Fu2020ASO}. In essence, KB in IR exploits the semantic regularities among data elements (e.g., concepts, documents), and researchers have recently leveraged deep learning to learn the regularities. For example, \citet{Lukovnikov2017NeuralNQ} trained an NN model to rank subject-predicate pairs to retrieve relevant facts for a given question. \citet{Shen2019MultiTaskLF} tackled the coreference resolution problem in semantic parsing, utilizing a large-scale knowledge base such as DBpedia~\citep{Lehmann2015DBpediaA} and Wikidata~\citep{Vrandei2014WikidataAF}.


KB involves in every aspect of ConvSearch. \citet{Agarwal2018AKM} incorporate a multi-modal knowledge base in generating system responses related to product-specific attributes, whereby a response can be a mixture of textual and visual information. \citet{Gaur2021ISEEQIS} utilize a knowledge graph to generate information-seeking questions. Clearly, KB can improve the NLP~\citep{Wang2019ContextualCD} and recommendation~\citep{Zhou2020ImprovingCR} modules in conversational search systems.

\subsection{Question-Answering}

The properties of QA and IR overlap with information-seeking perspectives. A subtle difference between them is that IR considers a set of relevant documents to answer the user's question. In contrast, QA provides more specific information to a question, such as factoids, a list of entities, definitions, relationships, etc., Recently, QA captured more attention due to the increased interest in conversational aspects of the information-seeking process. There are two types of approaches in QA: Machine Reading Comprehension (MRC)~\citep{Chen2017ReadingWT,Seo2017BidirectionalAF} and Knowledge Graph-based Question-Answering (KGQA)~\citep{Bast2015MoreAQ}. ConvQA follows the same approaches. In this paper, we highlight two aspects of QA in IR. Firstly, the research efforts to address the challenges in ConvQA, and secondly, Biomedical QA (BQA) and its applications.

In ConvQA, the user's questions are not in a full-fledged interrogative sentence. Inherently, search systems are demanded to resolve more complex linguistic problems of conversations~\citep{gao2019neural}. Query rewriting is an approach to replenish the partial view of a user's information need by identifying repeated or omitted mentions and replacing them with more specific words~\citep{Vakulenko2021QuestionRF}. \citet{Christmann2019LookBY} proposed a method to better understand users' incomplete questions in terms of fact-centric entities and predicates over a knowledge graph, automatically inferring missing or ambiguous pieces in a conversation. \citet{gao2019neural} proposed a generative model that maps incomplete utterances to logical forms, which can be executed on a knowledge base in searching for answers. \citet{Qu2019AttentiveHS} focused more on the positional information of conversation history. They proposed an MRC model with the history attention mechanism to demonstrate the importance of positional information of utterances in conversation modeling.

Our underlying hypothesis is that we can improve information retrieval quality by judicious use of QA systems to find items containing clues to answer user questions and queries. It is imperative to automatically construct and maintain domain-specific knowledge bases along with a corpus and exploit them in ConvSearch.

\subsection{Recommender Systems}

Recommender Systems (RecSys) and IR have similar goals; both return a list of relevant items. However, the goals are slightly different in that IR attempts to find the best item for the given query. In contrast, recommender systems allow exploration of the search space to find the available items from different perspectives. Like other information-seeking fields, RecSys started employing conversational approaches (ConvRecSys) that enable interactions to refine the understanding of users' preferences~\citep{Jannach2021ASO}. 

The most crucial task in RecSys is user modeling regardless of the methods: collaborative filtering or content-based filtering. Learning user preferences or degree of expertise should be a desired attribute of the conversational search systems. With user modeling, a system can provide tailored search results and recommendations. Indeed, this personalization can raise privacy and opacity concerns~\citep{Wang2018TowardPP} since users do not have direct access to the information collected by the system. Nevertheless, the techniques for ConvRecSys are gaining more attention due to the shared tasks with IR and the benefits of providing recommendations in the information-seeking process.

Recommendations can improve search experiences in various aspects: (1) The system can select high-quality items based on other measures beyond their content-based features. (2) Recommendations also provide intuitive explanations that help humans understand why certain items are recommended~\citep{Zhang2020ExplainableRA}. (3) Ultimately, recommendations provide multi-dimensional views of the targeted information. For example, scholarly document search engines (e.g., Semantic Scholar) recommend highly related citations to a research topic of interest. We can recommend items such as citation~\citep{Jeong2020ACC}, publication venue~\citep{Medvet2014PublicationVR}, authors to follow, and even queries. A query suggestion technique like ``People Also Ask'' recommends the relevant alternative queries that lead to a different set of search results.


\subsection{Evaluation}

Although the evaluation component is not shown in Figure~\ref{fig:cir}, the study of evaluation methodology in ConvSearch is vital and still needs to be explored to a full extent. ConvSearch differs from traditional IR systems in that it involves multi-turn interactions (e.g., sessions) using more likely well-formed natural language questions. Traditional evaluation methods (based on the Cranfield paradigm) focus on computing scores to a single query that essentially indicate how many relevant items are found and are positioned higher in a ranked list of results (e.g., precision, recall, and their variants).

There exists a line of work investigating the evaluation methods for interactive IR. Earlier, \citet{Carterette2015DynamicTC} proposed dynamic test collections to systematically evaluate interactiveness using click information and time spent reading documents. \citet{Jiang2015AutomaticOE} developed evaluation methods for voice-activated intelligent assistants in terms of user utility in speech recognition and intent classification. However, these approaches do not directly measure the performance of search-oriented conversational interactions. 

\citet{Zhang2017InformationRE} generalized IR evaluation metrics as search simulation using the notions of interaction reward and cost; for example, precision can be defined as the ratio of interaction reward and cost. \citet{Lipani2021HowAI} developed an offline evaluation framework of ConvSearch based on the idea of ``subtopics'' that models novelty and diversity in search and recommendation. Specifically, they defined an evaluation metric called Expected Conversation Satisfaction (ECS) which estimates conversation satisfaction over many simulated dialogues. 
Various attempts have been made to assess ConvSearch systems in multiple dimensions, including user intent prediction~\cite{Qu2019UserIP}, next question prediction~\cite{Yang2017NeuralMM}, user satisfaction prediction~\cite{Sun2022TrackingSS}, sub-goals prediction as appeared in the TREC Conversational Assistant Track~\cite{Dalton2020TRECC2}.

\section{Aims of Biomedical ConvSearch~\label{sec:bio}}

To an increasing extent, people utilize search engines to seek health advice~\citep{Cross2021SearchEV}; either to extract up-to-date clinical knowledge or to elicit self-diagnosis to their medical queries before reaching a doctor. Moreover, the journal published by \citet{Koman2020PhysiciansPO} concluded that the data-seeking agents assist the growing world but mentioned a few challenges likely, performance of accessing accurate information, bio-data integrity, and data miss-use. However, the conversational system acts as social technology, which figured as the significant approach for social distance clinical treatment in this recent COVID-19 pandemic \citep{Su2021TheRO}. In this scenario, ConvSearch can aid in the information-seeking process by providing means of iterative and interactive communication.

This section shares the latest development in conversational search, especially in the biomedical and healthcare domains. We highlight diverse and inclusive examples (beyond the IR perspectives) that portray the aims of ConvSearch in the biomedical domains.

\subsection{Semantic Information Retrieval}

Most of the users searching for biomedical information may not be able to provide precise scientific terms to express their information needs. Though the terminology has been given righteously, there would be an occurrence of vocabulary mismatch and semantic gaps.

As the biomedical knowledge to be expressed in a query becomes more complex and sophisticated, researchers propose various data structures and language standards. For example, researchers model semantics using subject-predicate-object triples in SemMedDB~\citep{Kilicoglu2012SemMedDBAP}, entity sets for search~\citep{Shen2018EntitySS}, a graph of contextualized concepts in SemEHR~\citep{Wu2018SemEHRAG}, dense vectors in the vector space model~\citep{RastegarMojarad2017SemanticIR}, or biological modeling languages such as BEL~\citep{Slater2014RecentAI}.

\citet{Kiesel2021MetaInformationIC} stressed the importance of meta-information in conversational search systems, which provides contextualized support in document retrieval. The meta-information concepts include document length, cognitive search intents, credibility, and extra-topical dimensions such as geographical and temporal information. Although it is a very early stage to assess the efficacy of conversational interfaces for semantic IR, \citet{Preininger2021DifferencesII} demonstrated that users prefer using a conversational agent to seek a certain type of information in a pharmacologic knowledge base, such as administration, intravenous compatibility, drug class, and pharmacokinetics.

\subsection{Knowledge Base of Biomedical Data}

To achieve Bio-Conversational search system, the knowledge base has to be related to the health base information. The required information can be gathered from various popular healthcare resources, Pubmed datasets, HealthData.gov, NICHD, National Library of Medicine datasets, and Hugging-Face datasets. Further, these datasets were molded into a structural manner to develop meta paths without any ambiguity or uncertainty to achieve QA framework. The construction of graph representations assists the database mechanism in developing a meta-path structural way to retrieve the necessary information. Constructing the structural information and improving the accuracy with semantic correlations between modalities is discussed by \citet{9836826} according to Attention-based Generative Adversarial Hashing) GAH model. In medical surveillance, for retrieving biological information Medical Knowledge Graph (MKG) gives a lift to BioConv systems in constructing structural knowledge. During the process of extracting information, to avoid noise and faulty outcomes of the data \citet{9669650} proposed a hybrid neural network model with multi-head attention enhanced with Graph Convolution Networks (GCN), which captures complex relations and illustrates the certain necessary meta-paths to extract bio-information.

\subsection{Biomedical Question-Answering and Conversational Agents}

Regarding biomedical information-seeking problems, the most actively studied fields are question-answering (BQA) and conversational agents (healthcare chatbots). One of the early BQA systems is \textit{AskHERMES}~\citep{Cao2011AskHERMESAO}, which allows physicians to write a question in natural language and navigate possible answer sentences to meet their information needs. BQA is an essential task for clinical decision support and personal health information seeking, which has been extensively studied. Interested readers can refer to ~\citet{Jin2021BiomedicalQA}'s survey, highlighting recent developments and approaches to BQA. 

Another popular adaptation is using conversational agents (i.e., chatbots) to support sociopsychological health interventions~\citep{Henschel2021WhatMA}. \citet{Kretzschmar2019CanYP} discussed the strength and limitations of using chatbots in mental health support. \citet{Kowatsch2021ConversationalAA} demonstrated the potential of chatbots as a social mediator between healthcare professionals, patients, and family members. Physicians perceive conversational agents as a practical and innovative tool when seeking information in general; However, physicians emphasize the significance of obtaining specific and trustworthy answers~\citep{Koman2020PhysiciansPO}. Developing systems with the properties of interpretability and privacy-preserving mechanisms is a crucial topic in biomedical ConvSearch~\citep{Brundage2020TowardTA,Alam2021ExaminingTE}.

\subsection{Personalization}

Information seekers with different backgrounds and degrees of domain knowledge may express different information needs with the same query. Search engines should comprehend the underlying search intents by understanding the user's preference in context. Personalization is a process of dynamically changing the information access and content, interface, and search functionality to tailor search outcomes for a user's information need. In a broader scope, personalization in healthcare includes personalized medicine~\citep{Stark2019ALR} and services like education and therapy~\citep{Ghanvatkar2019UserMF}. 



There exist many concerns regarding personalization in the information-seeking scenario. As pointed out by \citet{Kocaballi2019ThePO}, most ConvSearch systems with a personalization feature are implemented without theoretical or evidence-based support, which can cause patient safety, privacy, and fairness issues. Personalization in ConvSearch is also prone to amplifying bias~\citep{Gerritse2020BiasIC}, such as filtering results by information sources and reducing the diversity, which is a potential violation of the usability principles~\citep{Nielsen1994EnhancingTE}. 

\subsection{Trustworthiness and Privacy Concerns}

Recently, more attention has been directed to trustworthiness and privacy issues in deep learning technologies, such as the knowledge transfer mechanism behind the pre-trained language models (PLM). For example, word embeddings obtained from a publicly available PLM (e.g., BERT) can easily produce unwanted outcomes due to the unclear definition of similarity in the feature vector space. As \citet{Noh2021ImprovedBW} pointed out, PLMs may not be able to distinguish words in the same context, which can be a critical issue in natural language understanding/generation in biomedical applications. For example, words in antonym relationships (e.g., high/low blood pressure, opioid and nonopioid) or disease-symptom relationships (e.g., pharyngitis and sore throats) may have similar semantic representations.

To address this problem, a rapidly growing body of work focuses on analyzing the knowledge captured by PLM. A typical approach is probing via prompt engineering, which utilizes standardized biomedical vocabularies and ontologies. Commonly used benchmarks for probing biomedical knowledge include BioLAMA~\cite{Sung2021CanLM}  and MedLAMA~\cite{Meng2022RewirethenProbeAC}. Another focus is the privacy leakage issue with PLM. \citet{Nakamura2021KARTPL} examine the privacy risk of language models pretrained with a document collection containing sensitive personal information. Using deep learning techniques is inevitable in ConvSearch systems, requiring more research to examine the potential pitfalls of transferring knowledge from other information sources.
\begin{figure*}[tbh]
    \centering
    \includegraphics[width=.85\linewidth]{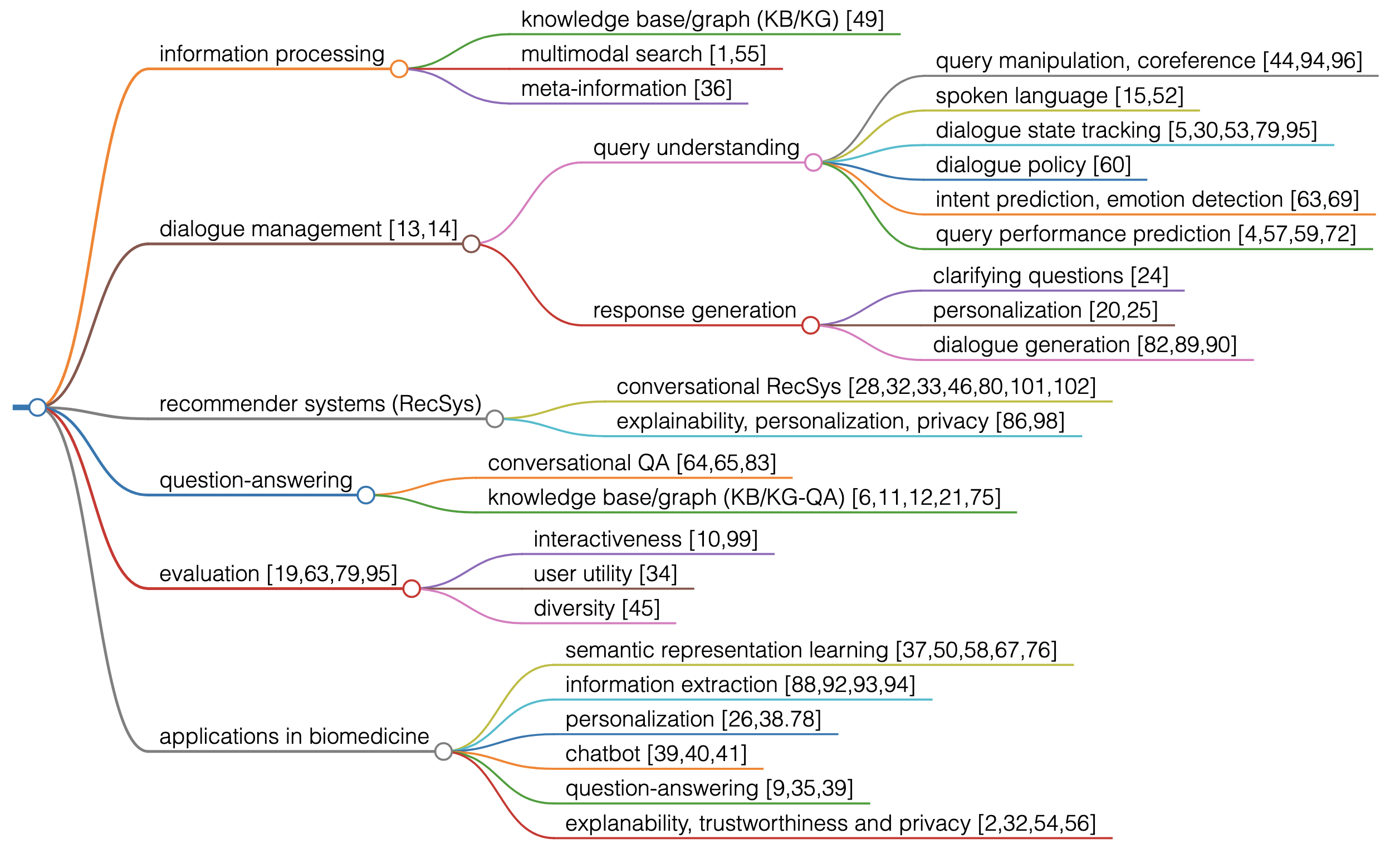}
    \caption{ConvSearch Research Sub-tasks}
    \label{fig:convsearch-thrusts}
\end{figure*}
\section{Conclusion}
Our survey summarizes current research on conversational search (ConvSearch). We have built our work on major components in ConvSearch and focused on the operation of dialogue and conversational interactions. Further, we have collaborated our work in the biomedical and healthcare domains to establish an ideology of social technology. The application of ConvSearch in the bio-field would bring a drastic change in the social world.

The emerging field of ConvSearch is adopting ideas from already established areas such as QA, RecSys, DS, and NLP, and the degree of integration is coarse at this stage (c.f., Fig.~\ref{fig:convsearch-thrusts}). For an instance, current ConvSearch research is heavily skewed towards QA; it should be supported by diverse set of language understanding techniques such as reasoning over knowledge derived from an information source utilizing knowledge bases. Consequently, policy learning acting as a `control tower' takes a vital role in the future development of ConvSearch systems. More researchers should focus on improving the integrated system of various methodologies from different fields. 
\bibliographystyle{ACM-Reference-Format}
\bibliography{sample-base}










\end{document}